\documentclass{scrartcl}
\pdfoutput=1
\usepackage[]{graphicx}\usepackage[]{color}
\makeatletter
\def\maxwidth{ %
  \ifdim\Gin@nat@width>\linewidth
    \linewidth
  \else
    \Gin@nat@width
  \fi
}
\makeatother

\definecolor{fgcolor}{rgb}{0.345, 0.345, 0.345}

\usepackage{framed}
\makeatletter
 {\par\unskip\endMakeFramed%
 \at@end@of@kframe}
\makeatother

\definecolor{shadecolor}{rgb}{.97, .97, .97}
\definecolor{messagecolor}{rgb}{0, 0, 0}
\definecolor{warningcolor}{rgb}{1, 0, 1}
\definecolor{errorcolor}{rgb}{1, 0, 0}
\newenvironment{knitrout}{}{} 

\usepackage{alltt}
\usepackage{latexsym}
\usepackage{graphicx}
\usepackage{mathptmx}
\usepackage{amsmath}
\usepackage{relsize}
\newif\ifpdffortikz
\pdffortikztrue
\ifpdffortikz
\else
\usepackage{tikz}
\usetikzlibrary{external}
\usetikzlibrary{automata,positioning, shapes,arrows,shadows,calc,matrix,decorations.pathmorphing}
\fi
\usepackage{subcaption} 
\usepackage[boxed]{algorithm2e}

\DeclareMathOperator{\argmax}{argmax}

\newcommand\CC{C\nolinebreak\hspace{-.05em}\raisebox{.4ex}{\relsize{-1}{\textbf{+}}}\nolinebreak\hspace{-.10em}\raisebox{.4ex}{\relsize{-1}{\textbf{+}}}}

\usepackage[utf8]{inputenc}
\usepackage{natbib} 
\IfFileExists{upquote.sty}{\usepackage{upquote}}{}
\begin{document}
%
%
%
%
\title{DATA-DRIVEN DYNAMIC DECISION MODELS\thanks{Copyright \textcopyright{} 2015, IEEE.}}%
%
\author{John J. Nay\\ [12pt]
School of Engineering\\
Vanderbilt University \\
PMB 351826 \\
2301 Vanderbilt Place\\
Nashville, TN 37235-1826, USA
\and
Jonathan M. Gilligan\\ [12pt]
Department of Earth \& Environmental Sciences\\
Vanderbilt University\\
PMB 351805\\
2301 Vanderbilt PL\\
Nashville, TN 37235-1805, USA
}
\maketitle

\section{ABSTRACT}
This article outlines a method for automatically generating models of dynamic decision-making that both have strong predictive power and are interpretable in human terms. This is useful for designing empirically grounded agent-based simulations and for gaining direct insight into observed dynamic processes. We use an efficient model representation and a genetic algorithm-based estimation process to generate simple approximations that explain most of the structure of complex stochastic processes. This method, implemented in \CC{} and \textsf{R}, scales well to large data sets. We apply our methods to empirical data from human subjects game experiments and international relations. We also demonstrate the method's ability to recover known data-generating processes by simulating data with agent-based models and correctly deriving the underlying decision models for multiple agent models and degrees of stochasticity.

\section{MODEL REPRESENTATION}
\label{sec:representation}

This article describes a modeling method designed to understand data on dynamic decision-making. We have created a practical, easy-to-use software package implementing the method. Although our method is more broadly applicable, the motivation for the model representation was prediction of individual behavior in strategic interactions, i.e.~games. Most \emph{behavioral} game-theoretic treatments of repeated games use action-learning models that specify the way in which attractions to actions are updated by an agent as play progresses \citep{camerer_behavioral_2003}. Action learning models can perform poorly at predicting behavior in games where cooperation (e.g.~Prisoner's Dilemma) or coordination (e.g.~Bach or Stravinsky) are key \citep{hanaki_action_2004}. Also, they often fail to account for the effects of changes in information and player matching conditions \citep{mckelvey_playing_2001}. In this paper, we model repeated game strategies as decision-making procedures that can explicitly consider the dynamic nature of the environment, e.g.~if my opponent cooperated last period then I will cooperate this period. We represent decision-making with finite-state machines and use a genetic algorithm to estimate the values of the state transition tables. This combination of representation and optimization allows us to efficiently and effectively model dynamic decision-making.

Traditional game theories define strategies as complete contingent plans that specify how a player will act in every possible state; however, when the environment becomes even moderately complex the number of possible states of the world can grow beyond the limits of human cognition \citep{miller_coevolution_1996,fudenberg_slow_2012}. One modeling response to cognitive limitations has been to exogenously restrict the complexity of repeated game strategies by representing them as Moore machines -- finite state machines whose outputs depend only on their current state \citep{moore_gedanken-experiments_1956} -- with a small number of states \citep{rubinstein_finite_1986,miller_coevolution_1996,hanaki_learning_2005}.
Moore machines can model bounded rationality, explicitly treating procedures of decision-making \citep{osborne_course_1994}. A machine modeling agent $i$ responding to the actions of agent $j$ is a four-tuple $[Q_{i}, q_{i}^0, f_{i}, \tau_{i}]$, where $Q_{i}$ is the set of states, $q_{i}^0 \in Q_{i}$ is the initial state, $f_{i}: Q_{i} \rightarrow A_{i}$ is the output function mapping a state to an action, and $\tau_{i}: Q_{i} \times A_{j} \rightarrow Q_{i}$ (where $j \neq i$) is the transition function mapping a state and another agent's action to a state \citep{osborne_course_1994}. We generalize this model beyond games by allowing for more inputs in $\tau_{i}$ than $A_{j}$, and by providing empirical rankings of these inputs that can be used to induce sparsity in more context-rich environments. The Moore machine can have many underlying states for a single observable action, allowing it to represent arbitrarily complex decision processes. The complexity is directly controlled by the number of states, which is a tuning parameter of our method that can be optimized by Algorithm~\ref{alg:cv} for predictive performance.

\begin{figure}[tb] 
\centering
\begin{subfigure}[b]{0.49\linewidth} \centering
\ifpdffortikz
    \includegraphics{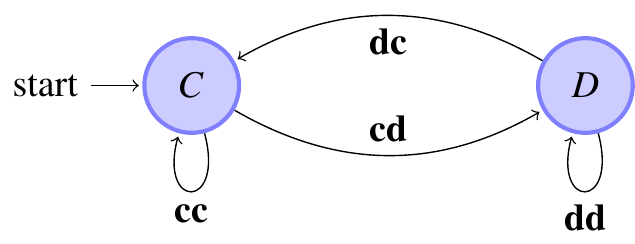}
\else
\begin{tikzpicture}[shorten >=1pt,node distance=4cm,on grid,auto]
\tikzstyle{every state}=[draw=blue!50,very thick,fill=blue!20]
   \node[state,initial] (C) {$C$};
   \node[state] (D) [right=of C] {$D$};
   \path[->]
    (C) edge [bend right] node {\textbf{cd}} (D)
        edge [loop below] node {\textbf{cc}} ()
    (D) edge [bend right] node {\textbf{dc}} (C)
        edge [loop below] node {\textbf{dd}} ();
\end{tikzpicture}
\fi
\caption{Tit-for-tat.}
\label{fig:automata-tft}
\end{subfigure}
\begin{subfigure}[b]{0.49\linewidth} \centering
\ifpdffortikz
\includegraphics{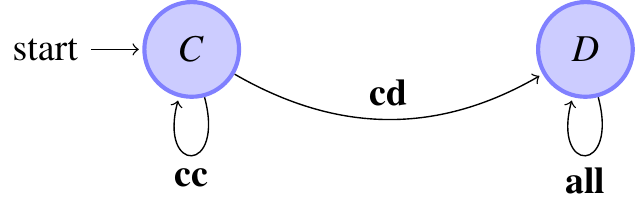}
\else
\begin{tikzpicture}[shorten >=1pt,node distance=4cm,on grid,auto]
\tikzstyle{every state}=[draw=blue!50,very thick,fill=blue!20]
   \node[state,initial] (C) {$C$};
   \node[state] (D) [right=of C] {$D$};
   \path[->]
    (C) edge [bend right] node {\textbf{cd}} (D)
        edge [loop below] node {\textbf{cc}} ()
    (D) edge [loop below] node {\textbf{all}} ();
\end{tikzpicture}
\fi
\caption{Grim trigger.}
\label{fig:automata-grim}
\end{subfigure}
\begin{subfigure}[b]{0.99\linewidth} \centering
\ifpdffortikz
\includegraphics{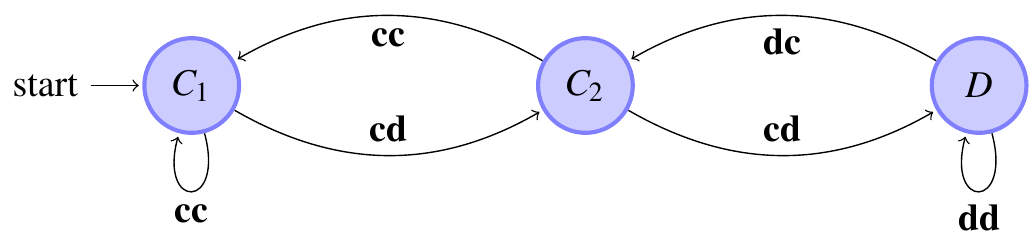}
\else
\begin{tikzpicture}[shorten >=1pt,node distance=4cm,on grid,auto]
\tikzstyle{every state}=[draw=blue!50,very thick,fill=blue!20]
\node[state,initial] (C1) {$C_1$};
\node[state] (C2) [right=of C1] {$C_2$};
\node[state] (D) [right=of C2] {$D$};
\path[->]
(C1) edge [bend right] node {\textbf{cd}} (C2)
edge [loop below] node {\textbf{cc}} ()
(C2) edge [bend right] node {\textbf{cd}} (D)
edge [bend right] node {\textbf{cc}} (C1)
(D) edge [bend right] node {\textbf{dc}} (C2)
edge [loop below] node {\textbf{dd}} ();
\end{tikzpicture}
\fi
\caption{Tit-for-two-tat.}
\label{fig:automata-tf2t}
\end{subfigure}
%
\caption[Game theoretic strategies represented as finite state machines.]{Game theoretic strategies represented as finite state machines \citep{rubinstein_finite_1986,miller_coevolution_1996,fudenberg_slow_2012}. Lower-case letters denote the possible outcomes of the previous period of play, e.g.~\textbf{cd} means the player cooperated and her opponent defected.}
\label{fig:automata}
\end{figure}
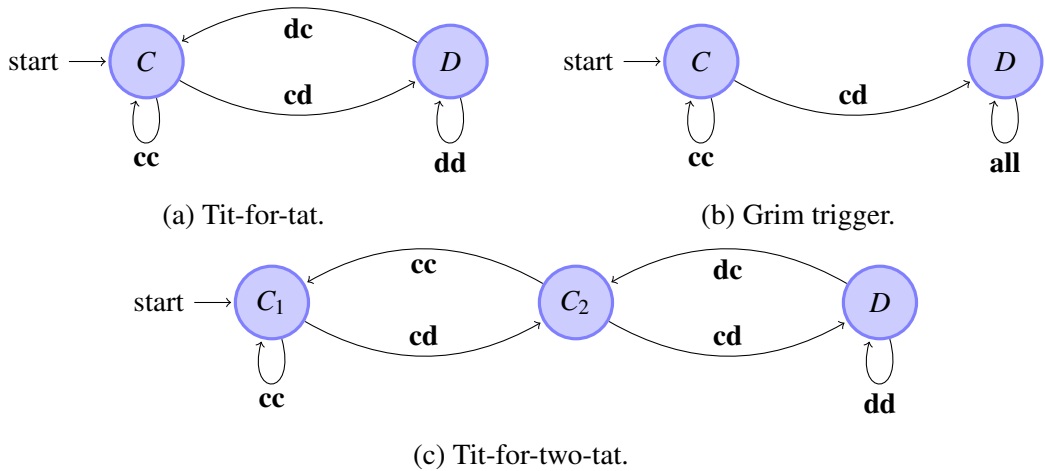

Fig.~\ref{fig:automata} shows examples of finite state machines (FSMs) representing strategies for the Iterated Prisoner's Dilemma game (see Section~\ref{sec:experimental} for game details): The possible states are cooperate ($C$) and defect ($D$), and after initialization the current state is determined by the history of the player and her opponent cooperating or defecting (\textbf{cc, cd, dc, dd}) in the previous period.

\section{ESTIMATION}
\label{sec:estimation}

Genetic algorithms (GAs) have been used to model agents updating beliefs based on endogenously determined variables in a general equilibrium environment \citep{bullard_using_1999}, and agents learning to make economic decisions \citep{arifovic_genetic_1994,arifovic_coordination_1995,marks_adaptive_1995,midgley_breeding_1997}. In contrast to investigations of GAs as models of agent learning and behavior, we use GAs to automatically generate interpretable agent decision models from empirical data. This is similar to work by \citet{fogel_ga_ipd_1993}, \citet{miller_coevolution_1996}, and \citet{miller_complex_2007}, in which GAs evolved FSMs based on their interactions with one another in simulated games, but whereas these were theoretical exercises, we are estimating models to explain and predict observed interactions among real agents. We use GAs as optimization routines for estimation because they perform well in rugged search spaces to quickly solve discrete optimization problems, are a natural complement to our binary string representation of FSMs \citep{goldberg_genetic_1988}, and are easily parallelized.

\citet{duffy_using_2002} combined empirical experimental data with genetic programming (GP) to model behavior. GP, with the same genetic operations as most GAs \citep{koza_genetic_1992}, is a process that can evolve arbitrary computer programs \citep{duffy_agent-based_2006}. We apply genetic operations to FSM representations rather than to all predictor variables and functional primitives because we are interested in deriving decision models with a particular structure: FSMs with latent states, rather than models conditioning on observable variables with any arbitrary functional form. With data-driven modeling, it is desirable to impose as many constraints as can be theoretically justified on the functional form of the model (see \citet{miller_complex_2007} for interesting theoretical results related to FSM agents interacting in games). This avoids overfitting by constraining the model to a functional form that is likely generalizable across contexts, allows genetic selection to converge better, and reduces the computational effort required to explore parameter space. An additional challenge in implementing GP is specifying the genetic operations on the function primitives while ensuring that they will always produce syntactically valid programs that represent meaningful decision models. This requires fine-tuning to specific problems, which we avoid because we are designing a general method applicable across domains.

Our choice to use Moore machines as the building blocks of our decision modeling method ensures that estimation will produce easily interpretable models with latent states that can be represented graphically (see Fig.~\ref{fig:automata} for examples). Our process represents Moore machines as Gray-encoded binary strings consisting of an action vector followed by elements that form the state matrix \cite{Savage_survey_1997}. For details, see Fig.~\ref{fig:bitstring} and our \verb+build_bitstring+, \verb+decode_action_vec+,  and \verb+decode_stat_mat+ functions. This way, genetic operators can have free reign to search the global parameter space guided by the ability to predict provided data with the decoded binary strings.

\begin{algorithm}
 \KwData{Time series of actions taken by agents and the relevant predictors of each action.}
 \KwResult{Finite state machine (FSM) with highest predictive performance.}
 Set convergence criteria (maximum number of generations or number of generations without improvement in performance of best FSM) based on the number of parameters to estimate\;
 Create initial population at step $k=0$ of $p$ individuals $\{ \theta_1^0,\theta_2^0,...,\theta_p^0 \}$ (FSMs, encoded as binary strings)\;
 \While{convergence not satisfied}{
  Decode each individual's string into an FSM, evaluate its predictive performance on training data, and set this as the individual's fitness for step $k$, $f(\theta_i^k)$\;
  Assign each individual a probability for selection proportional to its fitness, $p_i^k \propto f(\theta_i^k)$\;
  Select individuals by sampling from the population with replacement\;
  Create next generation, 
  $\{ \theta_1^{k+1},\theta_2^{k+1},...,\theta_p^{k+1} \}$, by applying random \emph{crossover\/} 
  and \emph{mutation\/}
  to the selected sub-population\;
 }
  Return $\argmax_{\theta_i^k}f(\theta_i^k)$: the individual with the greatest predictive power\;
  Check each element of the solution's state transition matrix for contribution to predictive performance, and evaluate solution on test data, if supplied.
 \caption{Evolving finite state machines with a genetic algorithm.}
 \label{alg:ga}
\end{algorithm}

The vast majority of computation time for Algorithm~\ref{alg:ga} is the evaluation of the predictive accuracy of the FSMs (not the stochastic generation of candidate FSMs). To improve performance we implement this evaluation in \CC\ using the \texttt{Rcpp} package \citep{eddelbuettel_rcpp_2013}, and, because it is embarrassingly parallel, distribute it across processor cores. We have incorporated our code into an \textsf{R} package with an API of documented function calls and using the \texttt{GA} package \citep{scrucca_ga_2013} to perform the GA evolution. A user can generate an optimized FSM by calling \verb+evolve_model(data)+, where \verb+data+ is an \textsf{R} \verb+data.frame+ object with columns representing the time period of the decision, the decision taken at that period, and any predictor variables. There are many additional optional arguments to this \verb+evolve_model+ function, but they have sensible default values. Our package then generates \CC\ code for a fitness function and uses it to evaluate automatically generated candidate models. Once the convergence criteria of this iterative search process is satisfied, the best FSM is identified, and each predictor variable is assessed by checking its identifiability and computing its importance in that decision model. The return value contains a descriptive summary of all results, including those shown in Fig.~\ref{fig:fsm-representations}.

%
%
%
%
The number of states in the FSM and the number of predictor variables to include are hyper-parameters that control the complexity of the model. Beginning with the simplest possible model and increasing complexity by adding states and variables, we often observe that at first, out-of-sample predictive accuracy grows because bias falls more quickly than variance rises; but eventually, adding further complexity reduces bias less than it increases variance so accuracy decreases \citep{hastie_elements_2009}. We can use cross-validation on the training data to find values for the hyper-parameters that maximize predictive accuracy (Algorithm~\ref{alg:cv}). We assess the out-of-sample predictive accuracy of the final model with a hold-out test set of data, distinct from the cross-validation test-sets in Algorithm~\ref{alg:cv}. Increasing complexity to optimize predictive accuracy introduces a new trade-off because more complex decision models are harder to interpret in human terms, so the ``best'' solution will depend on the goals of the analysis.

\begin{algorithm}
 \KwData{A dataset, a performance metric (e.g.~accuracy or area under the ROC curve), and possible hyper-parameter values (e.g.~number of states or predictor variables to include).}
 \KwResult{An estimate of the hyper-parameters that lead to best predictive performance.}
 \For{each row in design matrix of hyper-parameter sets}{
 Sample data into $k$ (e.g., 10) groups\; 
 \For{each $k$}{
 Set group $k$ as a testing set and everything else as a training set\;
 Evolve model on training set with Algorithm~\ref{alg:ga}\;
 Predict testing set and compare to actual values based on performance metric\;
 }
 Calculate average performance across all predictions on all $k$ testing sets\;
 }
 Return hyper-parameter set with best average performance\;
 \caption{Use cross-validation to optimize FSM hyper-parameters for predictive performance.}
 \label{alg:cv}
\end{algorithm}

\section{EXPERIMENTAL GAME DATA}
\label{sec:experimental}

\begin{figure}[tb]
\centering
\ifpdffortikz
\includegraphics{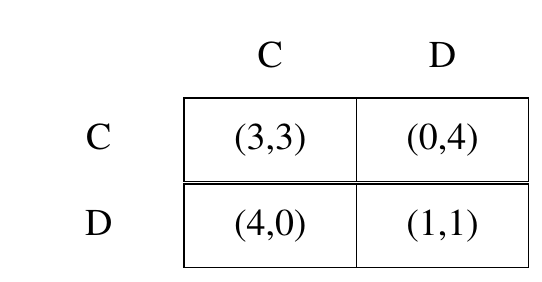}
\else
\begin{tikzpicture}[element/.style={minimum width=1.75cm,minimum height=0.85cm}]
\matrix (m) [matrix of nodes,nodes={element},column sep=-\pgflinewidth, row sep=-\pgflinewidth,]{
         & C  & D  \\
C & |[draw]|(3,3) & |[draw]|(0,4) \\
D & |[draw]|(4,0) & |[draw]|(1,1) \\
};
\end{tikzpicture}
\fi
\caption{Payoff table for a Prisoner's Dilemma game. C is cooperate and D is defect.}
\label{fig:payoff}
\end{figure}

The Iterated Prisoner's Dilemma (IPD) is often used as a model of cooperation \citep{axelrod_evolution_1984}. A one-shot PD game has a unique equilibrium in which each player chooses to defect even though both players would be better off if they cooperated. Suppose two players play the simultaneous-move PD game in Fig.~\ref{fig:payoff}, observe the choice of the other person, and then play the same simultaneous-move game again.
 Even in the (finitely) repeated version, no cooperation can be achieved by rational income maximizers. 
 This tension between maximizing collective and individual gain is representative of a broad class of social situations (e.g.~the ``tragedy of the commons'' \citep{hardin_tragedy_1968}). 

We applied our procedure to data from laboratory experiments on human subjects playing IPD games for real financial incentives. \citet{Nay_predicting_2014} gathered and integrated data from many experiments, conducted and analyzed by \citet{bereby-meyer_speed_2006,duffy_cooperative_2009,kunreuther_bayesian_2009,dal_bo_evolution_2011} and \citet{fudenberg_slow_2012}. All of the experiments share the same underlying repeated Prisoner's Dilemma structure, although the details of the games differed. Nay's data set comprises 135,388 cooperation decisions, which is much larger than previous studies of repeated game strategies. 

\citet{fudenberg_slow_2012} and \citet{dal_bo_evolution_2011} modeled their IPD experimental data with repeated game strategies; however, they applied a maximum likelihood estimation process to estimate the prevalence of a relatively small predefined set of strategies. In contrast, our estimation process automatically searches through a very large parameter space that includes all possible strategies up to a given number of states and does not require the analyst to predefine any strategies, or even understand the game.

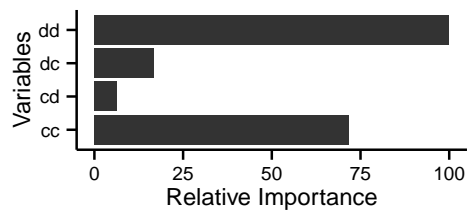
\begin{figure}[tb]
\centering
\begin{subfigure}[b]{0.46 \linewidth} \centering
\begin{tabular}{lrr}
raw string: & 
\multicolumn{2}{r}{
0, 1, 0, 0, 1, 0, 0, 0, 0, 0
} 
\\
encoded av, sm: & 
0, 1 &
0, 0, 1, 0, 0, 0, 0, 0
\\
decoded av, sm: & 
0, 1
&
0, 0, 1, 1, 1, 1, 1, 1
\end{tabular}
\caption{Raw string; string split into Gray-encoded action vector (av) and state matrix (sm); and decoded binary representations. Decoded sm consists of column-wise elements that index into the action vector to determine the 
action for each state.}
\label{fig:bitstring}
\end{subfigure}
\quad
\begin{subfigure}[b]{0.46 \linewidth} \centering
\begin{tabular}{lc}
Action vector: & 
1, 2 \\[2.5mm]
State matrix: &
\begin{tabular}{c|cccc}
 \textbf{State} & cc & dc & cd & dd \\
 \hline
 \textbf{C} (1) & 
 \textbf{1} & 
 \textit{2} & 
 \textbf{2} & 
 \textit{2} \\
 \textbf{D} (2) &
 \textit{1} & 
 \textbf{2}  & 
 \textit{2} &  
 \textbf{2} \\     
\end{tabular}
\end{tabular}

\caption{Bitstring decoded into action vector and state matrix. For this action vector, 1 corresponds to cooperation (C) and 2 to defection (D). Columns of state matrix correspond to the observed behaviors at $t-1$ and rows correspond to the state.}
\label{fig:decoded}
\end{subfigure}
\begin{subfigure}[b]{0.46 \linewidth} \centering
\ifpdffortikz
\includegraphics{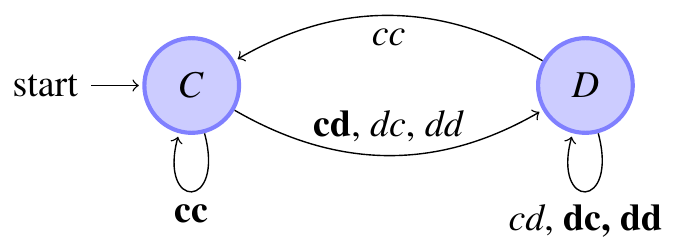}
\else
\begin{tikzpicture}[shorten >=1pt,node distance=4cm,on grid,auto]
\tikzstyle{every state}=[draw=blue!50,very thick,fill=blue!20]
   \node[state,initial] (C) {$C$};
   \node[state] (D) [right=of C] {$D$};
   \path[->]
    (C) edge [bend right] node {\textbf{cd}, \textit{dc}, \textit{dd}} (D)
        edge [loop below] node {\textbf{cc}} ()
    (D) edge [bend right] node {\textit{cc}} (C)
    (D) edge [loop below] node {\textit{cd}, \textbf{dc, dd}} ();
\end{tikzpicture}
\fi
\caption{Corresponding graph representation.}
\label{fig:decoded-fsm}
\end{subfigure}
\quad
\begin{subfigure}[b]{0.46 \linewidth} \centering
\begin{knitrout}
\definecolor{shadecolor}{rgb}{0.969, 0.969, 0.969}\color{fgcolor}
\includegraphics[width=\maxwidth]{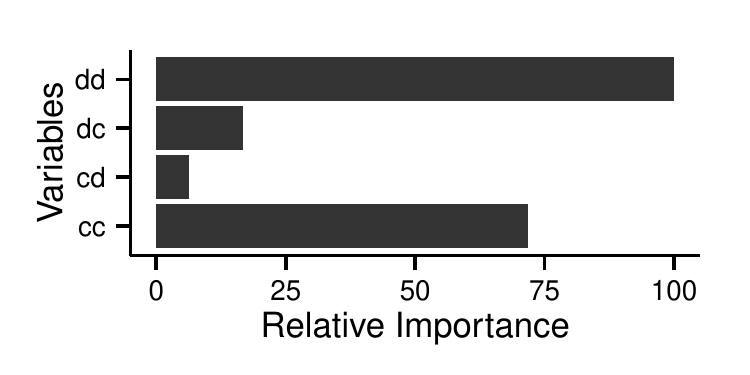} 

\end{knitrout}

\caption{Relative variable importance. Most important is 100.}
\label{fig:varImp_experimental_data}
\end{subfigure}

\caption[Empirically estimated finite state machine.]{A finite state machine estimated on 108,305~decisions with 82\% accuracy on a hold-out test data set of 27,083~decisions.
Transitions that would be accessible in strictly deterministic play are represented in boldface and inaccessible transitions in italic. Because the human players did not follow exact deterministic strategies, and the italicized transitions were taken in simulating this model with actual game play, the values of these transitions were identifiable.
}
\label{fig:fsm-representations}
\end{figure}

We used 80\% of our data for training and reserved the other 20\% as a hold-out test set. Fig.~\ref{fig:fsm-representations} shows different representations of the the fittest two-state machine of a GA population evolved on the training data: The raw Gray-encoded and binary string (Fig.~\ref{fig:bitstring}), the bitstring decoded into state matrix and action vector form (Fig.~\ref{fig:decoded}), and the corresponding graph representation (Fig.~\ref{fig:decoded-fsm}). We measure variable importance (Fig.~\ref{fig:varImp_experimental_data}) by switching each value of an estimated model's state matrix to another value in its feasible range, measuring the decrease in goodness of fit to the training data, normalizing the values, then summing across each column to estimate the relative importance of each predictor variable (in this case, the moves each player made in the previous turn).

\begin{figure}[tb]
\centering
\begin{knitrout}
\definecolor{shadecolor}{rgb}{0.969, 0.969, 0.969}\color{fgcolor}
\includegraphics[width=\maxwidth]{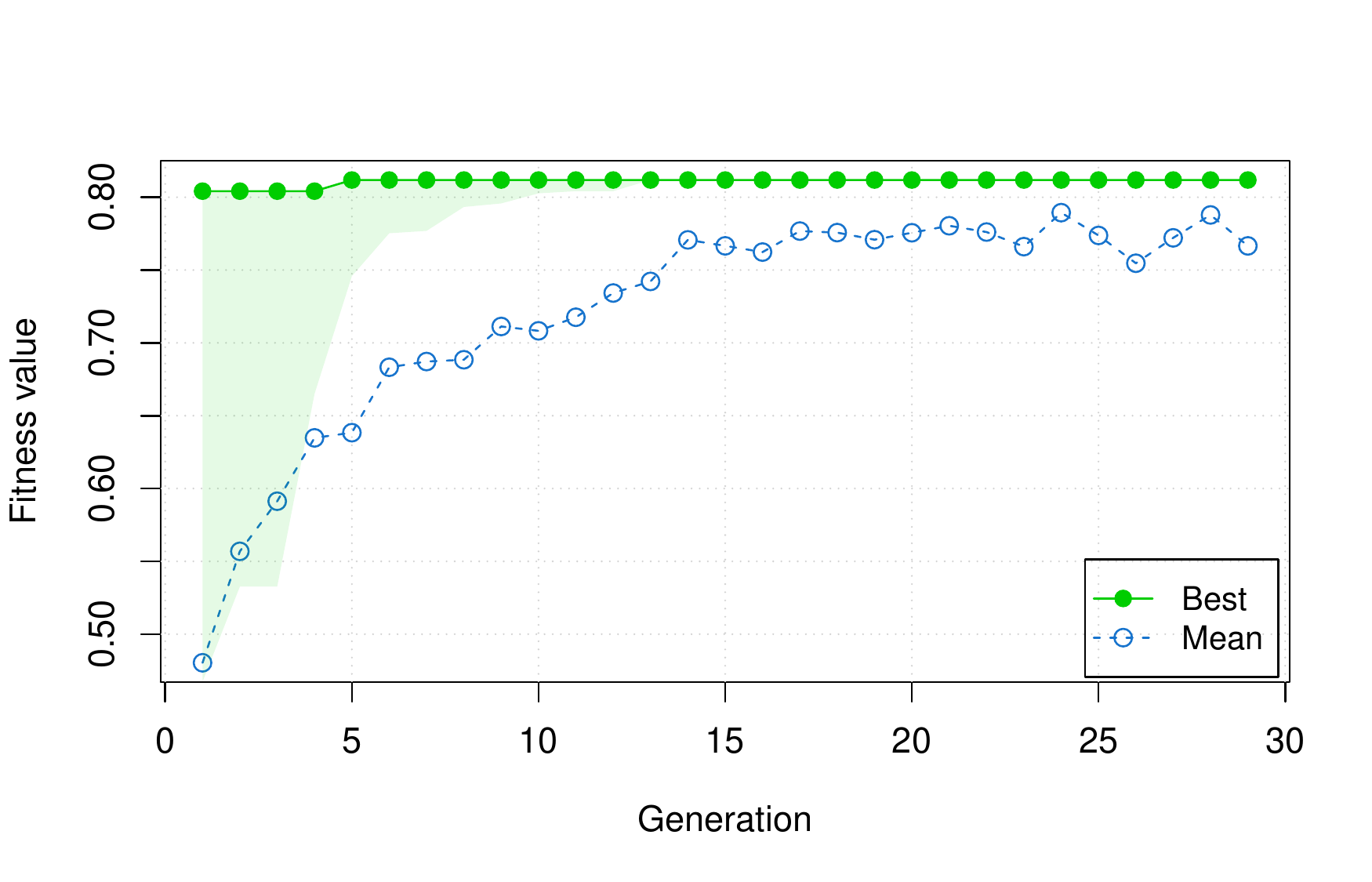} 

\end{knitrout}
\caption[GA predictions.]{Proportion of correct predictions of human decisions in IPD games by an evolving population of decision models (strategies). On average, a random choice strategy will have 50\% accuracy. Shaded area is the difference between the median of the population of strategies and the best individual strategy.}
\label{fig:ga_fsm}
\end{figure}

Fig.~\ref{fig:ga_fsm} illustrates the GA run that evolved the FSM of Fig.~\ref{fig:fsm-representations} by predicting cooperation decisions in IPD training data games. This GA run, which only took a few seconds on a modest laptop, used common algorithm settings: a population of 175 FSMs initialized with random bitstrings. If the analyst has an informed prior belief about the subjects' decision models, she can initialize the population with samples drawn from that prior distribution, but this paper focuses on deriving useful results from random initializations, corresponding to uniform priors, where the analyst only provides data. A linear-rank selection process used the predictive ability of individuals to select a subset of the population from which to create the next generation. A single-point crossover process was applied to the binary values of selected individuals with 0.8 probability, uniform random mutation was conducted with probability 0.1, and the top 5\% fittest individuals survived each generation without crossover or mutation, ensuring that potentially very good solutions would not be lost \citep{scrucca_ga_2013}. These are standard GA parameter settings and can be adjusted if convergence is taking particularly long for a given dataset. 


Using theoretical agent-based simulations and a fitness measure that is a function of simulated payoffs, \citet{axelrod_complexity_1997} demonstrated the fitness of the tit-for-tat (TFT) strategy. Using a fitness measure that is a function of the ability to explain human behavior, we discovered a hybrid of TFT and grim trigger (GT), which we call noisy grim (NG). TFT's state is determined solely by the opponent's last play. GT will never exit a defecting state, no matter what the opponent does.

With traditional repeated game strategies such as TFT and GT, the player always takes the action corresponding to her current state (boldface transitions in Fig.~\ref{fig:decoded-fsm}), but if we add noise to decisions so the player will sometimes choose the opposite action from her current state (italic transitions in Fig.~\ref{fig:decoded-fsm}), then the possibility arises for both the player and opponent to cooperate when the player is in the defecting state (i.e.~to reach the second row first column position of the state matrix in Fig.~\ref{fig:decoded}). This would return the player to the cooperating state (see, e.g., \citet{chong_noisy_ipd_2005}). Noisy grim's predictions on the hold-out test data are 82\% accurate, GT's are 72\% accurate, and TFT's are 77\% accurate. We also tested 16 other repeated game strategies for the IPD from \citep{fudenberg_slow_2012}. Their accuracy on the test set ranged from 46\% to 77\%. Our method uncovered a deterministic dynamic decision model that predicts IPD play better than all of the existing theoretical automata models of IPD play that we are aware of and has interesting relationships to the two most well-known models: TFT and GT.


This process has allowed us to estimate a highly interpretable decision model (fully represented by the small image of Fig.~\ref{fig:decoded-fsm}) that predicts most of the behavior of hundreds of human participants, merely by plugging in the dataset as input. We address the potential concern that the process is too tuned to this specific case study by inputting a very different dataset from the field of international relations and obtaining useful results. However, before moving to more empirical data---where the data-generating process can never be fully known---to test how robustly we can estimate a known model, we repeatedly simulate a variety of known data-generating mechanisms and then apply the method to the resulting choice data.

\section{SIMULATED DATA}
\label{sec:simulated}

In the real world, people rarely strategically interact by strictly following a deterministic strategy \citep{chong_noisy_ipd_2005}. Whimsy, strategic randomization, or error may induce a player to choose a different move from the one dictated by her strategy. To study whether our method could determine an underlying strategy that an agent would override from time to time, we followed the approach of \citet{fudenberg_slow_2012} and created an agent-based model of the IPD in which agents followed deterministic strategies, such as TFT and GT, but made noisy decisions: At each time period, the deterministic strategy dictates each agent's preferred action, but the agent will choose the opposite action with probability $p$, where $p$ ranges from 0 (perfectly deterministic play) to 0.5 (completely random play). The noise parameter, $p$, is constant across all states of a strategy of a particular agent for any given simulation experiment we conducted.

When a player follows an unknown strategy, characterized by latent states, discovering the strategy (the actions corresponding to each state and transitions between the states) requires observed data that explores as much as possible of the state transition matrix defined by all possible combinations of state and predictor values (for these strategies the predictors are the history of play). Many deterministic strategy pairings can quickly reach equilibria in which players repeat the same moves for the rest of the interaction. If the player and opponent both use TFT and both make the same first move, every subsequent move will repeat the first move. If the opponent plays GT, then after the first time the player defects the opponent will defect for the rest of the session and the data will provide little information on the player's response to cooperation by the opponent. However, if the opponent plays with noise, the play will include many instances of cooperation and defection by the opponent, and will thus sample the accessible state space for the player's strategy more thoroughly than if the opponent plays deterministically. Indeed, this is why \citet{fudenberg_slow_2012} added noise to action choices in their human subjects experimental games.

\begin{figure}[tbp]
\centering
\begin{subfigure}[b]{\linewidth}\centering
\begin{knitrout}
\definecolor{shadecolor}{rgb}{0.969, 0.969, 0.969}\color{fgcolor}
\includegraphics[width=\maxwidth]{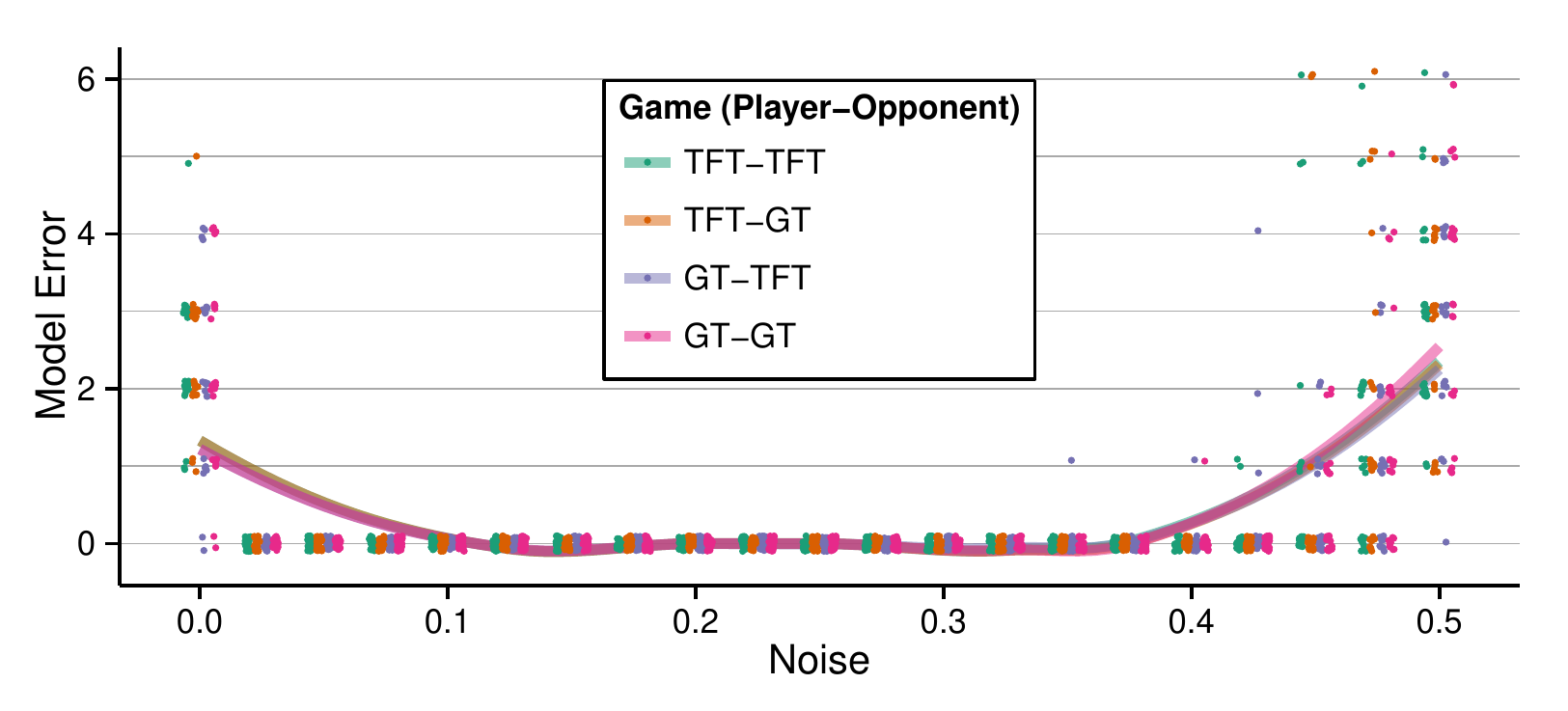} 

\end{knitrout}
\caption[ ]{Effect of decision models and noise in choice on ability to estimate true model when a noisy player faces an equally noisy opponent. Color indicates the pair of decision models used by the player and opponent. For noise probabilities between 0.025 and 0.325 the estimated model always exactly matches the true model.\label{fig:game-choice-noise}}
\end{subfigure}


\begin{subfigure}[b]{\linewidth}\centering
\begin{knitrout}
\definecolor{shadecolor}{rgb}{0.969, 0.969, 0.969}\color{fgcolor}
\includegraphics[width=\maxwidth]{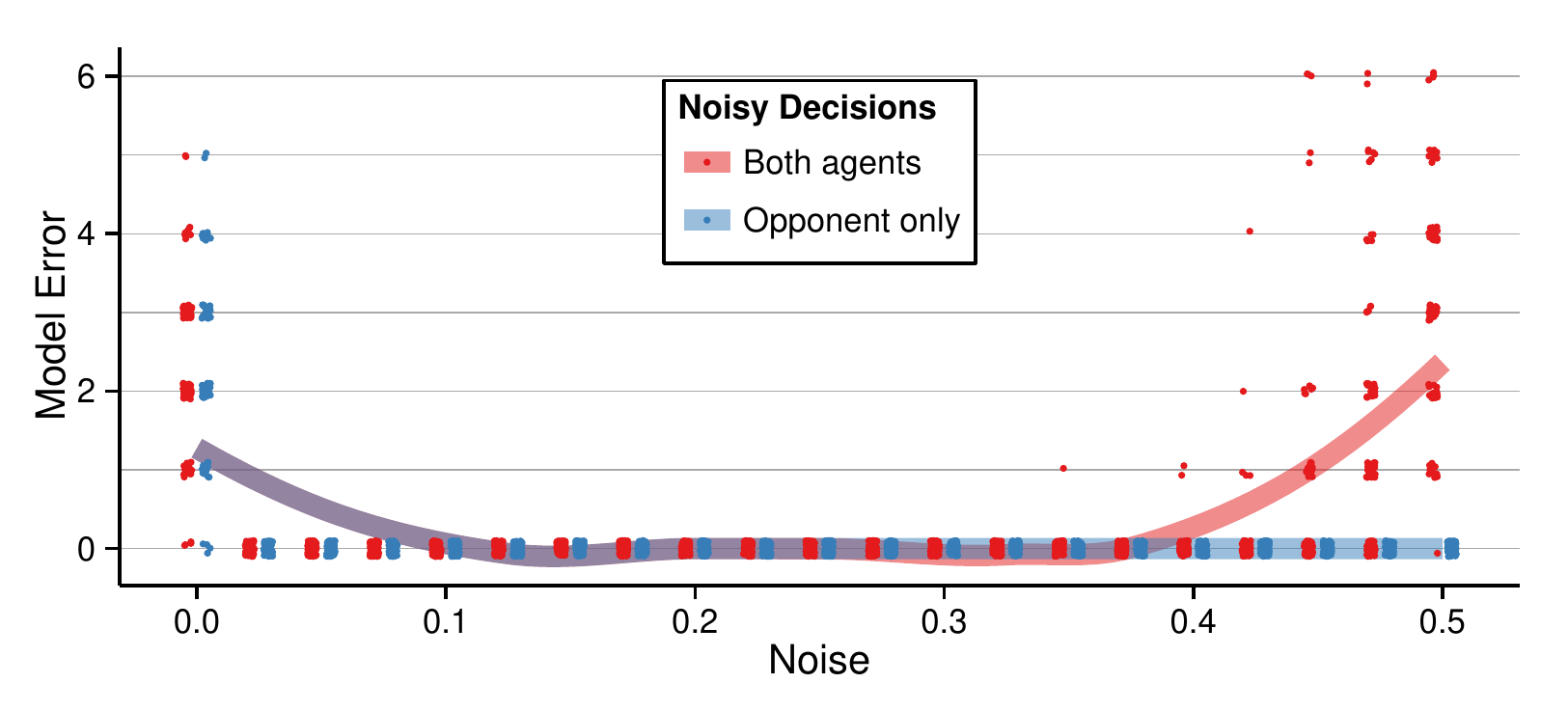} 

\end{knitrout}
\caption[ ]{Comparison of a purely deterministic player facing a noisy opponent versus a noisy player facing a noisy opponent. As noise approaches 0.5, the decision model's impact on actual choices vanishes, so the model error rises sharply when the player's decisions are noisy.\label{fig:player-opponent-noise}}
\end{subfigure}

\caption[ ]{Role of noise in correctly identifying the player's strategy. Points have been jittered for visualization. Lines are LOESS-curves. Noise represents the probability, in any period, that an agent will take the opposite action from its decision model.
Both the GT and TFT strategies can be represented as a 10 element matrix, of which 6 elements have identifiable values. The model error represents the number of mismatched elements between the estimated and true matrices, so it ranges from 0 (best) to 6 (worst).}
\label{fig:exp_results}
\end{figure}

We simulated approximately 17 million interactions, varying paired decision models of each agent [(\emph{TFT}, \emph{TFT\/}), (\emph{TFT}, \emph{GT\/}), (\emph{GT}, \emph{TFT\/}), (\emph{GT}, \emph{GT\/})] and also varying the noise parameter $p$ (0, 0.025, ... , 0.5) for each of two noise conditions: where both players made equally noisy decisions, and where only the opponent made noisy decisions while the player under study strictly followed a deterministic strategy. We ran 25 replicates of each of the 168~experimental conditions, with 4,000 iterations of game play for each replicate, and then applied the FSM estimation method to each replicate of the simulated choice data to estimate the strategy that the agent player under study was using.

Being in state/row $k$ (e.g.~2) corresponds to the player taking action $k$ (e.g.~D) in the current turn. All entries in row $k$ corresponding to the player taking action $k$ in the current period (e.g.~columns 2 and 4 for D) are identifiable. Entries in row $k$ that correspond to not taking action $k$ in the current period (e.g.~columns 1 and 3 for row 2) represent transitions that cannot occur in strictly deterministic play, so their values cannot affect play and thus cannot be determined empirically. We take this into account when testing the method's ability to estimate underlying deterministic models: this is why only 6 elements of a 10-element TFT or GT matrix can be identified (Fig.~\ref{fig:exp_results}). We also take this into account when estimating models from empirical data, where the data-generating process is assumed to be stochastic: each element of the matrix that would be inaccessible under deterministic play is identified, and the fitness is calculated with a strategy matrix in which that element is changed to its complement (``flipped''). If flipping the element does not change the fitness, then the two complementary strategies are indistinguishable and the element in question cannot be determined empirically. If each element decreases the fitness when it is flipped,  then the strategy corresponds to a deterministic approximation of a stochastic process and all of the elements of the state matrix can be identified.

When the noise parameter was zero, most of the models estimated by the GA had at least two incorrect elements. However, for moderate amounts of noise ($p = 0.025$--$0.325$), all of the models estimated by the GA were correct (see Fig.~\ref{fig:game-choice-noise}). For noise levels above $p = 0.325$ in the player, the amount of error rose rapidly with $p$, as expected because at $p = 0.5$ the action the player chooses moves completely at random so there is no strategy to discover. When a strictly deterministic player faced a noisy opponent, the GA correctly identified the player's strategy for all noise levels above $p = 0.025$ (see Fig.~\ref{fig:player-opponent-noise}).

\section{OBSERVATIONAL DATA}
\label{sec:observational}

In order to extend this method to more complex situations the predictor variables (columns of the state matrices) can include any time-varying variable relevant to an agent's decision. In context-free games such as the IPD, the only predictor variables are the moves the players made in the previous turn, but models of strategic interactions in context-rich environments may include other relevant variables.

We find it difficult to interpret graphical models with more than four predictors, but an analyst who had many potentially relevant predictor variables and was unable to use theory alone to reduce the number of predictors sufficiently to generate easily interpretable models with our method could take four courses of action (listed in order of increasing reliability and computation time):
\begin{enumerate}
  \item Before FSM estimation, apply a (multivariate or univariate) predictor variable selection method. 
  \item Before FSM estimation, estimate an arbitrary predictive model that can produce variable importance rankings and then use the top $p<4$ predictors for FSM estimation. 
  \item After FSM estimation with $p \ge 4$ predictors, inspect the returned predictor variable importance ranking, and remove all but the top $p<4$ from her dataset and re-run estimation. 
  \item Conduct FSM estimation with all combinations of $p<4$ predictors out of all relevant predictors and choose the estimated model with the best performance (usually highest out-of-sample accuracy).
\end{enumerate}

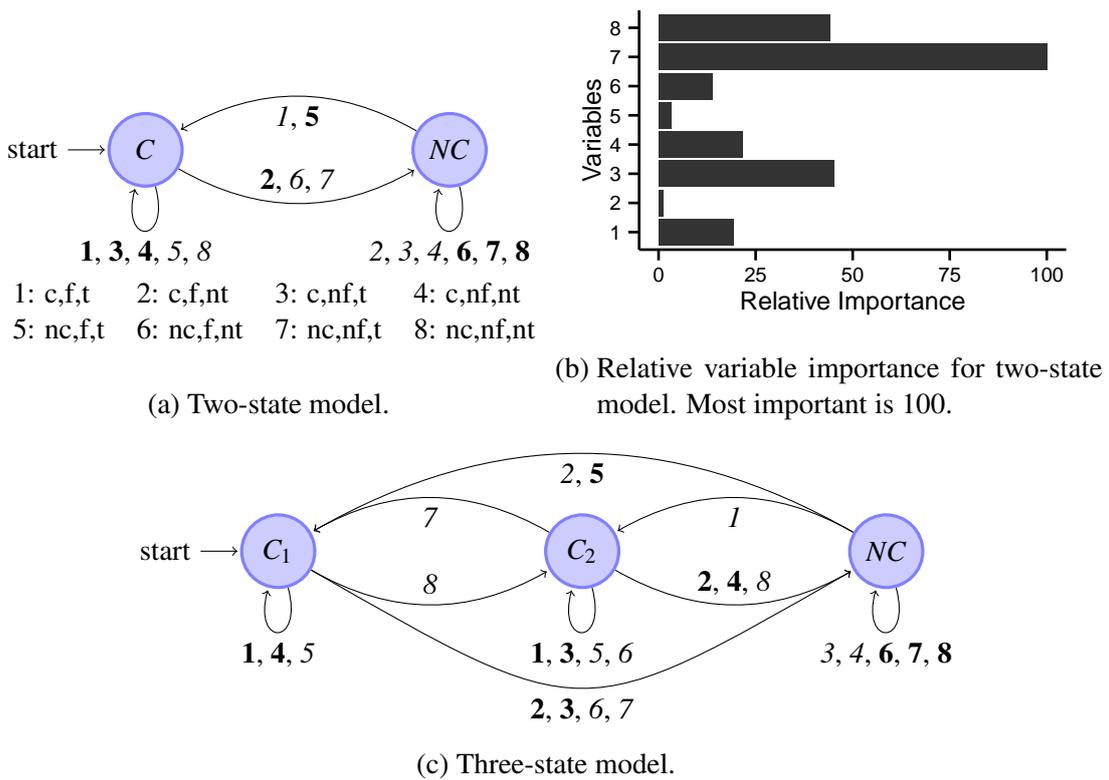
\begin{figure}[tb]
\centering
\begin{subfigure}[b]{0.5\linewidth}\centering
\ifpdffortikz
\includegraphics{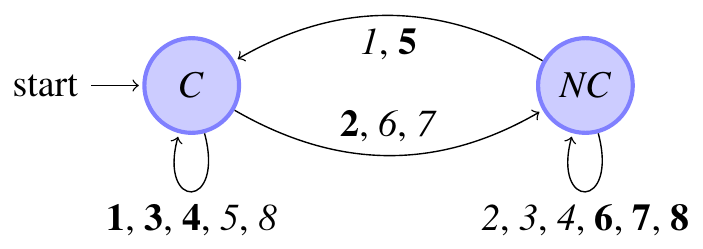}
\else
\begin{tikzpicture}[shorten >=1pt,node distance=4cm,on grid,auto]
\tikzstyle{every state}=[draw=blue!50,very thick,fill=blue!20]
   \node[state,initial] (C) {$C$};
   \node[state] (D) [right=of C] {$NC$};
   \path[->]
    (C) edge [bend right] node {\textbf{2}, \textit{6}, \textit{7}} (D)
        edge [loop below] node {\textbf{1}, \textbf{3}, \textbf{4}, \textit{5}, \textit{8}} ()
    (D) edge [bend right] node {\textit{1}, \textbf{5}} (C)
        edge [loop below] node {\textit{2}, \textit{3}, \textit{4}, \textbf{6}, \textbf{7}, \textbf{8}} ();
\end{tikzpicture}
\fi
\vspace{3mm}
\scalebox{1.0}{
\begin{tabular}{lllll}
1:  {c},{f},{t} & 
2:  {c},{f},{nt} & 
3:  {c},{nf},{t} &
4:  {c},{nf},{nt} \\
5: {nc},{f},{t} & 
6: {nc},{f},{nt} & 
7: {nc},{nf},{t} & 
8: {nc},{nf},{nt}
\end{tabular}
}
\caption[Two-state model.]{Two-state model.}
\label{fig:obs-two-state}
\end{subfigure}
\begin{subfigure}[b]{0.49\linewidth} \centering
\begin{knitrout}
\definecolor{shadecolor}{rgb}{0.969, 0.969, 0.969}\color{fgcolor}
\includegraphics[width=\maxwidth]{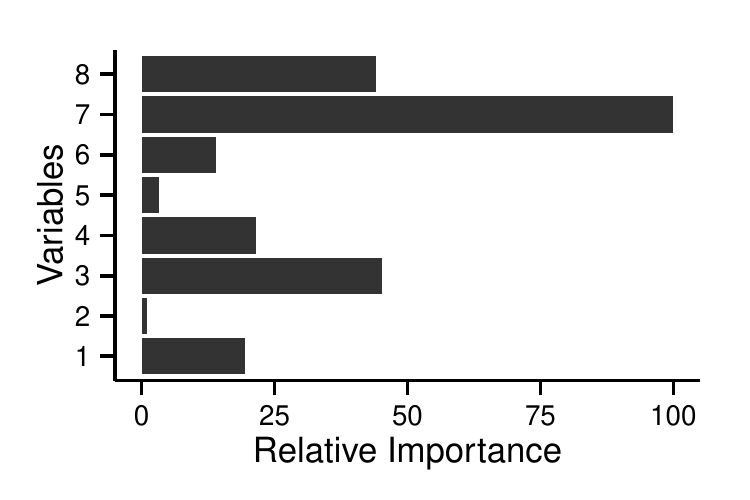} 

\end{knitrout}
\caption{Relative variable importance for two-state model. Most important is 100.}
\label{fig:varImp_obs_data}
\end{subfigure}
\begin{subfigure}[b]{\linewidth}\centering
\ifpdffortikz
\includegraphics{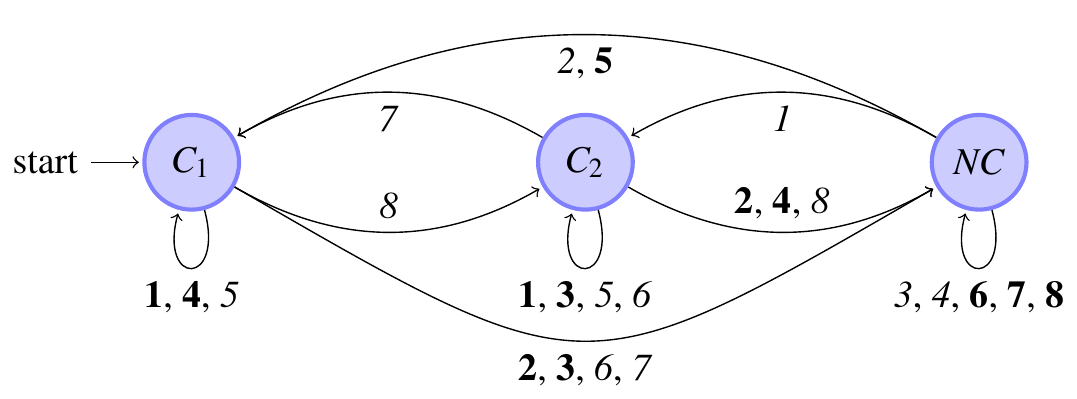}
\else    
\begin{tikzpicture}[shorten >=1pt,node distance=4cm,on grid,auto]
\tikzstyle{every state}=[draw=blue!50,very thick,fill=blue!20]
   \node[state,initial] (C1) {$C_1$};
   \node[state] (C2) [right=of C1] {$C_2$};
   \node[state] (D) [right=of C2] {$NC$};
   \path[->]
    (C1) edge [bend right] node {\textit{8}} (C2)
         edge [bend right, looseness=1.5] node [swap] {\textbf{2}, \textbf{3}, \textit{6}, \textit{7}} (D)
        edge [loop below] node {\textbf{1}, \textbf{4}, \textit{5}} ()
    (C2) edge [bend right] node {\textbf{2}, \textbf{4}, \textit{8}} (D)
         edge [bend right] node {\textit{7}} (C1)
        edge [loop below] node {\textbf{1}, \textbf{3}, \textit{5}, \textit{6}} ()
    (D) edge [bend right] node {\textit{2}, \textbf{5}} (C1)
        edge [bend right] node {\textit{1}} (C2)
        edge [loop below] node {\textit{3}, \textit{4}, \textbf{6}, \textbf{7}, \textbf{8}} ();
\end{tikzpicture}
\fi
\caption[]{Three-state model.}
\label{fig:obs-three-state}
\end{subfigure}
\caption{Models of cooperation over water issues for sets of countries. The italicized transitions would not be feasible outcomes if this strategy was played deterministically, but they are realized in the data. The inputs 
(c,nc) represent cooperation and non-cooperation, respectively; (f,nf) conflict and non-conflict; and (t,nt) signing and not signing a treaty.}
\label{fig:observational}
\end{figure}

We illustrate the use of extra predictor variables by applying our method to an example from international relations involving repeated water management-related interactions between countries that share rivers. We use data compiled by \citet{brochmann_signing_2012} on treaty signing and cooperation over water quality, water quantity, and flood control from 1948--1999 to generate a model for predicting whether two countries will cooperate. We used three lagged variables: whether there was water-related conflict between them in the previous year, whether they cooperated around water in the previous year, and whether they had signed a water-related treaty during any previous year. This data set was too small to divide into training and hold-out subsets for assessing predictive accuracy, so we report models' accuracy in reproducing the training data (a random choice model is 50\% accurate). A two-state decision model (Fig.~\ref{fig:obs-two-state}) is 73\% accurate, a three-state model (Fig.~\ref{fig:obs-three-state}) is 78\% accurate, and a four-state model is 82\% accurate, but its complexity makes it difficult to interpret visually so it is not shown.

Accuracy can be a problematic measure when the classes are imbalanced, i.e.~if a class the model is trying to predict is rare. Many alternatives to accuracy are available that illuminate different aspects of predictive power. For instance, precision is the proportion of (cooperation) event signals predicted by our models that are correct and recall is the proportion of events that are predicted by our models. For this subset of the dataset, cooperate and not cooperate were almost evenly distributed and to maintain a comparison to the experimental and simulated data we used accuracy as the fitness measure.

In the two-state model, whether or not the countries cooperated in the previous year, the combination of conflict and treaty-signing in the previous year always produces cooperation, whereas conflict without treaty-signing in the previous year always produces non-cooperation. In the three-state model, three of the four outcomes that include conflict lead to a transition from non-cooperation to cooperation, and four of the six outcomes that cause transitions from cooperation (states $C_1$ and $C_2$) to non-cooperation are non-conflict outcomes. While this does not tell us something decisive about the role of conflict, it suggests that there may be a counter-intuitive role of conflict in promoting cooperation. \citet{brochmann_signing_2012}, using a bivariate probit simultaneous equation model, has a similar finding: ``In the aftermath of conflict, states may be particularly eager to solve important issues that could cause future problems'' (p.~159).

\section{DISCUSSION}
This paper outlined a method for estimating interpretable models of dynamic decision-making. By estimating a global, deterministic, simple function for a given dataset, imposing constraints on the number of predictor variables, and providing options for reducing the number of predictor variables, our process facilitates capturing a significant amount of information in a compact and useful form. The method can be used for designing empirically grounded agent models in agent-based simulations and for gaining direct insight into observed behaviors of real agents in social and physical systems. Combining state matrices and a genetic algorithm has proven effective for simulated data, experimental game data, and observational international relations data. With the simulated data, we successfully recovered the exact underlying models that generated the data. With the real data, we estimated simple deterministic approximations that explain most of the structure of the unknown underlying process. We discovered a theoretically interesting dynamic decision model that predicted IPD play better than all of the existing theoretical models of IPD play that we were aware of.


We have released \href{https://github.com/JohnNay/datafsm}{an open-source \textsf{R} package} that implements the methods described here to estimate any time series classification model that uses a small number of binary predictor variables and moves back and forth between the values of the outcome variable over time. Larger sets of predictor variables can be reduced to smaller sets by applying one of the four methods outlined in Section~\ref{sec:observational}. Although the predictor variables must be binary, a quantitative variable can be converted into binary by division of the observed values into high/low classes. Future releases of the package may include additional estimation methods to complement GA optimization.

\section*{ACKNOWLEDGMENTS}

We gratefully acknowledge the authors of \href{http://www.r-project.org}{\textsf{R}} \citep{r_core}. 
This manuscript was prepared using 
knitr \citep{xie_knitr_2014}. We would like to thank Yevgeniy Vorobeychik for discussions on predicting games. 

This work was supported by U.S. National Science Foundation grants EAR-1416964 and EAR-1204685.

\bibliography{ga-fsm-wsc}

\begin{thebibliography}{36}
\providecommand{\natexlab}[1]{#1}
\providecommand{\url}[1]{\texttt{#1}}
\expandafter\ifx\csname urlstyle\endcsname\relax
  \providecommand{\doi}[1]{doi: #1}\else
  \providecommand{\doi}{doi: \begingroup \urlstyle{rm}\Url}\fi

\bibitem[Arifovic(1994)]{arifovic_genetic_1994}
Jasmina Arifovic.
\newblock Genetic algorithm learning and the cobweb model.
\newblock \emph{Journal of Economic Dynamics and Control}, 18:\penalty0 3--28,
  1994.

\bibitem[Arifovic and Eaton(1995)]{arifovic_coordination_1995}
Jasmina Arifovic and Curtis Eaton.
\newblock Coordination via genetic learning.
\newblock \emph{Computational Economics}, 8:\penalty0 181--203, 1995.
\newblock \doi{10.1007/BF01298459}.

\bibitem[Axelrod(1984)]{axelrod_evolution_1984}
Robert~M. Axelrod.
\newblock \emph{The Evolution of Cooperation}.
\newblock Basic Books, New York, 1984.

\bibitem[Axelrod(1997)]{axelrod_complexity_1997}
Robert~M. Axelrod.
\newblock \emph{The Complexity of Cooperation: Agent-based Models of
  Competition and Collaboration}.
\newblock Princeton University Press, Princeton, 1997.

\bibitem[Bereby-Meyer and Roth(2006)]{bereby-meyer_speed_2006}
Yoella Bereby-Meyer and Alvin~E. Roth.
\newblock The speed of learning in noisy games: Partial reinforcement and the
  sustainability of cooperation.
\newblock \emph{The American Economic Review}, 96:\penalty0 1029--1042, 2006.

\bibitem[Brochmann(2012)]{brochmann_signing_2012}
Marit Brochmann.
\newblock Signing river treaties: Does it improve river cooperation?
\newblock \emph{International Interactions}, 38:\penalty0 141--163, 2012.
\newblock \doi{10.1080/03050629.2012.657575}.

\bibitem[Bullard and Duffy(1999)]{bullard_using_1999}
James Bullard and John Duffy.
\newblock Using genetic algorithms to model the evolution of heterogeneous
  beliefs.
\newblock \emph{Computational Economics}, 13:\penalty0 41--60, 1999.
\newblock \doi{10.1023/A:1008610307810}.

\bibitem[Camerer(2003)]{camerer_behavioral_2003}
Colin~F. Camerer.
\newblock \emph{Behavioral Game Theory: Experiments in Strategic Interaction}.
\newblock Princeton University Press, Princeton, 2003.

\bibitem[Chong and Yao(2005)]{chong_noisy_ipd_2005}
S.Y. Chong and Xin Yao.
\newblock Behavioral diversity, choices and noise in the iterated prisoner's
  dilemma.
\newblock \emph{IEEE Transactions on Evolutionary Computation}, 9:\penalty0
  540--551, 2005.

\bibitem[Dal~Bo and Frechette(2011)]{dal_bo_evolution_2011}
Pedro Dal~Bo and Guillaume~R Frechette.
\newblock The evolution of cooperation in infinitely repeated games:
  Experimental evidence.
\newblock \emph{American Economic Review}, 101:\penalty0 411--429, 2011.
\newblock \doi{10.1257/aer.101.1.411}.

\bibitem[Duffy(2006)]{duffy_agent-based_2006}
John Duffy.
\newblock Agent-based models and human subject experiments.
\newblock In \emph{Handbook of Computational Economics}, volume~2, pages
  949--1011. Elsevier, Amsterdam, 2006.

\bibitem[Duffy and Engle-Warnick(2002)]{duffy_using_2002}
John Duffy and Jim Engle-Warnick.
\newblock Using symbolic regression to infer strategies from experimental data.
\newblock In \emph{Evolutionary Computation in Economics and Finance}, pages
  61--82. Springer, New York, 2002.

\bibitem[Duffy and Ochs(2009)]{duffy_cooperative_2009}
John Duffy and Jack Ochs.
\newblock Cooperative behavior and the frequency of social interaction.
\newblock \emph{Games and Economic Behavior}, 66:\penalty0 785--812, 2009.
\newblock \doi{10.1016/j.geb.2008.07.003}.

\bibitem[Eddelbuettel(2013)]{eddelbuettel_rcpp_2013}
Dirk Eddelbuettel.
\newblock \emph{Seamless {R} and {\CC} Integration with {Rcpp}}.
\newblock Springer, New York, 2013.

\bibitem[Fogel(1993)]{fogel_ga_ipd_1993}
David~B. Fogel.
\newblock Evolving behaviors in the iterated prisoner's dilemma.
\newblock \emph{Evolutionary Computation}, 1:\penalty0 77--97, 1993.

\bibitem[Fudenberg et~al.(2012)Fudenberg, Rand, and
  Dreber]{fudenberg_slow_2012}
Drew Fudenberg, David~G Rand, and Anna Dreber.
\newblock Slow to anger and fast to forgive: Cooperation in an uncertain world.
\newblock \emph{American Economic Review}, 102:\penalty0 720--749, 2012.
\newblock \doi{10.1257/aer.102.2.720}.

\bibitem[Goldberg and Holland(1988)]{goldberg_genetic_1988}
David~E. Goldberg and John~H. Holland.
\newblock Genetic algorithms and machine learning.
\newblock \emph{Machine Learning}, 3:\penalty0 95--99, 1988.
\newblock \doi{10.1023/A:1022602019183}.

\bibitem[Hanaki(2004)]{hanaki_action_2004}
Nobuyuki Hanaki.
\newblock Action learning versus strategy learning.
\newblock \emph{Complexity}, 9:\penalty0 41--50, 2004.

\bibitem[Hanaki et~al.(2005)Hanaki, Sethi, Erev, and
  Peterhansl]{hanaki_learning_2005}
Nobuyuki Hanaki, Rajiv Sethi, Ido Erev, and Alexander Peterhansl.
\newblock Learning strategies.
\newblock \emph{Journal of Economic Behavior \& Organization}, 56:\penalty0
  523--542, 2005.
\newblock \doi{10.1016/j.jebo.2003.12.004}.

\bibitem[Hardin(1968)]{hardin_tragedy_1968}
Garrett Hardin.
\newblock The tragedy of the commons.
\newblock \emph{Science}, 162:\penalty0 1243--1248, 1968.
\newblock \doi{DOI: 10.1126/science.162.3859.1243}.

\bibitem[Hastie et~al.(2009)Hastie, Tibshirani, and
  Friedman]{hastie_elements_2009}
Trevor Hastie, Robert Tibshirani, and Jerome Friedman.
\newblock \emph{The {Elements} of {Statistical} {Learning}: {Data} {Mining},
  {Inference}, and {Prediction}, {Second} {Edition}}.
\newblock Springer, New York, NY, 2nd edition, 2009.
\newblock ISBN 9780387848570.

\bibitem[Koza(1992)]{koza_genetic_1992}
John~R. Koza.
\newblock \emph{Genetic Programming: On the Programming of Computers by Means
  of Natural Selection}.
\newblock Bradford, Cambridge, MA, 1992.

\bibitem[Kunreuther et~al.(2009)Kunreuther, Silvasi, Bradlow, and
  Small]{kunreuther_bayesian_2009}
Howard Kunreuther, Gabriel Silvasi, Eric~T. Bradlow, and Dylan Small.
\newblock Bayesian analysis of deterministic and stochastic prisoner's dilemma
  games.
\newblock \emph{Judgment and Decision Making}, 4:\penalty0 363--384, 2009.

\bibitem[Marks et~al.(1995)Marks, Midgley, and Cooper]{marks_adaptive_1995}
Robert~E. Marks, David~F. Midgley, and Lee~G. Cooper.
\newblock Adaptive behaviour in an oligopoly.
\newblock In J\"org Biethahn and Volker Nissen, editors, \emph{Evolutionary
  Algorithms in Management Applications}, pages 225--239. Springer, New York,
  1995.

\bibitem[McKelvey and Palfrey(2001)]{mckelvey_playing_2001}
Richard~D. McKelvey and Thomas~R. Palfrey.
\newblock Playing in the dark: Information, learning, and coordination in
  repeated games.
\newblock Technical report, California Institute of Technology, Pasadena, 2001.

\bibitem[Midgley et~al.(1997)Midgley, Marks, and Cooper]{midgley_breeding_1997}
David~F. Midgley, Robert~E. Marks, and Lee~C. Cooper.
\newblock Breeding competitive strategies.
\newblock \emph{Management Science}, 43:\penalty0 257--275, 1997.
\newblock \doi{10.1287/mnsc.43.3.257}.

\bibitem[Miller(1996)]{miller_coevolution_1996}
John~H. Miller.
\newblock The coevolution of automata in the repeated prisoner's dilemma.
\newblock \emph{Journal of Economic Behavior \& Organization}, 29:\penalty0
  87--112, 1996.

\bibitem[Miller and Page(2007)]{miller_complex_2007}
John~H. Miller and Scott~E. Page.
\newblock \emph{Complex Adaptive Systems: An Introduction to Computational
  Models of Social Life}.
\newblock Princeton University Press, Princeton, 2007.

\bibitem[Moore(1956)]{moore_gedanken-experiments_1956}
Edward Moore.
\newblock Gedanken-experiments on sequential machines.
\newblock \emph{Automata Studies}, 34:\penalty0 129--153, 1956.

\bibitem[Nay(2014)]{Nay_predicting_2014}
John~Jacob Nay.
\newblock Predicting cooperation and designing institutions: An integration of
  behavioral data, machine learning, and simulation.
\newblock In \emph{Winter Simulation Conference Proceedings}, Savannah, {GA},
  December 2014.

\bibitem[Osborne and Rubinstein(1994)]{osborne_course_1994}
Martin~J. Osborne and Ariel Rubinstein.
\newblock \emph{A Course in Game Theory}.
\newblock {MIT} Press, Cambridge, MA, 1994.

\bibitem[{R Core Team}(2014)]{r_core}
{R Core Team}.
\newblock \emph{R: A Language and Environment for Statistical Computing}.
\newblock R Foundation for Statistical Computing, Vienna, Austria, 2014.
\newblock URL \url{http://www.R-project.org/}.

\bibitem[Rubinstein(1986)]{rubinstein_finite_1986}
Ariel Rubinstein.
\newblock Finite automata play the repeated prisoner's dilemma.
\newblock \emph{Journal of Economic Theory}, 39\penalty0 (1):\penalty0 83--96,
  June 1986.
\newblock ISSN 0022-0531.
\newblock \doi{10.1016/0022-0531(86)90021-9}.
\newblock URL
  \url{http://www.sciencedirect.com/science/article/pii/0022053186900219}.

\bibitem[Savage(1997)]{Savage_survey_1997}
C.~Savage.
\newblock A {Survey} of {Combinatorial} {Gray} {Codes}.
\newblock \emph{SIAM Review}, 39\penalty0 (4):\penalty0 605--629, January 1997.
\newblock ISSN 0036-1445.
\newblock \doi{10.1137/S0036144595295272}.
\newblock URL \url{http://epubs.siam.org/doi/abs/10.1137/S0036144595295272}.

\bibitem[Scrucca(2013)]{scrucca_ga_2013}
Luca Scrucca.
\newblock {GA}: A package for genetic algorithms in {R}.
\newblock \emph{Journal of Statistical Software}, 53:\penalty0 1--37, 2013.
\newblock URL \url{http://www.jstatsoft.org/v53/i04/}.

\bibitem[Xie(2014)]{xie_knitr_2014}
Yihui Xie.
\newblock \emph{Dynamic Documents with R and knitr}.
\newblock Chapman \& Hall/CRC, Boca Raton, 2014.

\end{thebibliography}

\end{document}